\documentclass[acmsmall, nonacm]{acmart}
\usepackage{ascmac}
\usepackage{booktabs}
\usepackage{multirow}
\usepackage{physics}

\newcommand{\soone}{\ensuremath{\text{S}_\text{Order1}}}
\newcommand{\sotwo}{\ensuremath{\text{S}_\text{Order2}}}
\newcommand{\scone}{\ensuremath{\text{S}_\text{Comp1}}}
\newcommand{\sctwo}{\ensuremath{\text{S}_\text{Comp2}}}
\newcommand{\sego}{\ensuremath{\text{S}_\text{Egoc}}}
\newcommand{\ssal}{\ensuremath{\text{S}_\text{Verb}}}
\newcommand{\sband}{\ensuremath{\text{S}_\text{Band}}}
\newcommand{\sattn}{\ensuremath{\text{S}_\text{Attn}}}

\begin{document}
\title{Cognitive Biases in Large Language Models: A Survey and Mitigation Experiments}
\titlenote{The extended abstract of this paper is presented at the 40th ACM/SIGAPP Symposium on Applied Computing (SAC 2025)}

\author{Yasuaki Sumita}
\affiliation{
  \institution{Kyoto University}
  \city{Kyoto} 
  \country{Japan}
}
\email{ysumita@ml.ist.i.kyoto-u.ac.jp}

\author{Koh Takeuchi}
\affiliation{
  \institution{Kyoto University}
  \city{Kyoto}
  \country{Japan}
}
\email{takeuchi@i.kyoto-u.ac.jp}

\author{Hisashi Kashima}
\affiliation{
  \institution{Kyoto University}
  \city{Kyoto} 
  \country{Japan}
}
\email{kashima@i.kyoto-u.ac.jp}

\begin{abstract}
Large Language Models (LLMs) are trained on large corpora written by humans and demonstrate high performance on various tasks. However, as humans are susceptible to cognitive biases, which can result in irrational judgments, LLMs can also be influenced by these biases, leading to irrational decision-making. For example, changing the order of options in multiple-choice questions affects the performance of LLMs due to order bias. In our research, we first conducted an extensive survey of existing studies examining LLMs' cognitive biases and their mitigation. The mitigation techniques in LLMs have the disadvantage that they are limited in the type of biases they can apply or require lengthy inputs or outputs. We then examined the effectiveness of two mitigation methods for humans, SoPro and AwaRe, when applied to LLMs, inspired by studies in crowdsourcing. To test the effectiveness of these methods, we conducted experiments on GPT-3.5 and GPT-4 to evaluate the influence of six biases on the outputs before and after applying these methods. The results demonstrate that while SoPro has little effect, AwaRe enables LLMs to mitigate the effect of these biases and make more rational responses.
\end{abstract}



\maketitle

\section{Introduction}\label{sec-intro}

In recent years, Large Language Models (LLMs) such as GPT-4 \cite{openai2023gpt4} have been developed rapidly and have shown high performance on various tasks such as machine translation \cite{kocmi2023large}, summarization \cite{stiennon2020learning}, and annotation \cite{gilardi2023chatgpt}.
This high level of performance is achieved by learning from large corpora of human-written documents.

However, humans exhibit various cognitive biases that lead to irrational decisions.
Cognitive bias is a systematic pattern of deviation from norm or rationality in judgment \cite{tversky1974judgment}.
For example, humans often adjust their estimates insufficiently away from initial values.
This is called ``anchoring'' \cite{tversky1974judgment}.
Another example is the ``bandwagon effect'', a tendency to adopt certain behaviors, styles, or attitudes simply because others are doing so \cite{leibenstein1950bandwagon}.

Since these cognitive biases also affect human-written documents, LLMs can inherit these biases during training, leading to irrational outputs.
For example, in multiple-choice questions, changing the order of options affects the performance of LLMs due to order bias \cite{pezeshkpour2024large}.
Additionally, when LLMs generate answers to questions, including irrelevant information in the question text can lead them to provide incorrect answers due to anchoring \cite{jones2022capturing}.

Several studies work on mitigating these cognitive biases in LLMs.
For example, to deal with order bias, some methods have been proposed to generate multiple outputs for the same question by changing the order of the options \cite{hou2024large} or to output the reasons along with the answers \cite{zheng2023judging}.
However, these methods have limited applicability, or it is unclear if they are also applicable to other biases.
Additionally, These methods require longer inputs and outputs or asking LLMs the same questions many times.

In this study, we first survey previous studies that discuss cognitive biases in LLMs and how to mitigate them (Table \ref{tabl:cb-works}).
LLMs show various types of cognitive biases.
Conversely, existing mitigation methods have the disadvantage that they are either limited in the type of biases they can apply or require lengthy inputs or outputs.

In addition, we experimentally investigated the application of the two cognitive bias mitigation methods used in crowdsourcing to prompt input into LLMs.
One method is SoPro (Social Projection), which prompts LLMs to answer as they believe other people would.
The other is AwaRe (Awareness Reminder), which makes LLMs aware of biases and encourages careful responses.
Since these methods only involve changing prompts, LLMs do not need to process the same questions repeatedly or generate lengthy responses.
Furthermore, these methods can be applied to various cognitive biases, as they are not constrained by the format of the questions.

To evaluate our methods, we conducted experiments on GPT-3.5 and GPT-4 using CoBBLEr (Cognitive Bias Benchmark for Large Language Models as Evaluators) \cite{koo2024benchmarking} to compare the effect of six cognitive biases before and after application.
The results indicate that while SoPro is ineffective, AwaRe encourages more rational responses and successfully mitigates these biases.
The result of SoPro applications is inconsistent with that for humans.

{\tabcolsep=0.1em
	\begin{table*}[t]
		\centering
		\caption{A summary of the cognitive biases discussed in the related works. It illustrates the types of cognitive biases considered in each study and whether their mitigation is discussed.}
		\label{tabl:cb-works}
		\resizebox{\textwidth}{!}{
			\begin{tabular}{c|ccccccccccccccccccccccccccccccccccccccc}
				\hline
				{Reference}                    & \cite{zhao2021calibrate} & \cite{sinclair2022structural} & \cite{binz2023using} & \cite{jones2022capturing} & \cite{lampinen2024language} & \cite{shi2023large} & \cite{talboy2023challenging} & \cite{suri2024large} & \cite{bian2024influence} & \cite{hou2024large} & \cite{wang2024large} & \cite{zheng2023judging} & \cite{wu2023style} & \cite{itzhak2024instructed} & \cite{pezeshkpour2024large} & \cite{shaki2023cognitive} & \cite{zheng2023large} & \cite{ranaldi2024hans} & \cite{koo2024benchmarking} & \cite{saito2023verbosity} & \cite{wang2023primacy} & \cite{tjuatja2024llms} & \cite{ma2023large} & \cite{liu2024selfsupervised} & \cite{opedal2024language} & \cite{eicher2024reducing} & \cite{berberette2024redefining} & \cite{schmidgall2024addressing} & \cite{macmillan-scott2024irrationality} & \cite{echterhoff2024cognitive} \\ \hline
				{Mitigation}                   & \checkmark               &                               &                      &                           &                             & \checkmark          &                              &                      &                          & \checkmark          & \checkmark           & \checkmark              &                    &                             & \checkmark                  &                           & \checkmark            & \checkmark             &                            &                           &                        &                        & \checkmark         & \checkmark                   &                           & \checkmark                    &                                 & \checkmark                      &                                         & \checkmark                     \\ \hline
				{Acquiescence bias}            &                          &                               &                      &                           &                             &                     &                              &                      &                          &                     &                      &                         &                    &                             &                             &                           &                       &                        &                            &                           &                        & \checkmark             &                    &                              &                           &                               &                                 &                                 &                                         &                                \\ \hline
				{Anchoring}                    &                          &                               &                      & \checkmark                &                             &                     & \checkmark                   & \checkmark           &                          &                     &                      &                         &                    &                             &                             & \checkmark                &                       &                        &                            &                           &                        &                        &                    &                              &                           &                               &                                 &                                 &                                         & \checkmark                     \\ \hline
				{Attentional bias}             &                          &                               &                      &                           &                             & \checkmark          &                              &                      &                          &                     &                      &                         &                    &                             &                             &                           &                       &                        & \checkmark                 &                           &                        &                        &                    &                              &                           &                               &                                 &                                 &                                         &                                \\ \hline
				{Attribute substitution}       &                          &                               &                      & \checkmark                &                             &                     &                              &                      &                          &                     &                      &                         &                    &                             &                             &                           &                       &                        &                            &                           &                        &                        &                    &                              &                           &                               &                                 &                                 &                                         &                                \\ \hline
				{Authority bias}               &                          &                               &                      &                           &                             &                     &                              &                      & \checkmark               &                     &                      &                         &                    &                             &                             &                           &                       &                        &                            &                           &                        &                        &                    &                              &                           &                               &                                 &                                 &                                         &                                \\ \hline
				{Availability bias}            & \checkmark               &                               &                      & \checkmark                &                             &                     &                              & \checkmark           &                          &                     &                      &                         &                    &                             &                             &                           &                       &                        &                            &                           &                        &                        &                    &                              &                           &                               & \checkmark                      & \checkmark                      &                                         &                                \\ \hline
				{Bandwagon effect}             &                          &                               &                      &                           &                             &                     &                              &                      &                          &                     &                      &                         &                    &                             &                             &                           &                       &                        & \checkmark                 &                           &                        &                        &                    &                              &                           &                               &                                 & \checkmark                      &                                         &                                \\ \hline
				{Base rate neglect}            &                          &                               &                      &                           &                             &                     & \checkmark                   &                      &                          &                     &                      &                         &                    &                             &                             &                           &                       &                        &                            &                           &                        &                        &                    &                              &                           &                               &                                 &                                 &                                         &                                \\ \hline
				{Belief bias}                  &                          &                               &                      &                           &                             &                     &                              &                      &                          &                     &                      &                         &                    & \checkmark                  &                             &                           &                       &                        &                            &                           &                        &                        &                    &                              &                           &                               &                                 &                                 &                                         &                                \\ \hline
				{Certainty effect}             &                          &                               & \checkmark           &                           &                             &                     &                              &                      &                          &                     &                      &                         &                    & \checkmark                  &                             &                           &                       &                        &                            &                           &                        &                        &                    &                              &                           &                               &                                 &                                 &                                         &                                \\ \hline
				{Compassion fade}              &                          &                               &                      &                           &                             &                     &                              &                      &                          &                     &                      &                         &                    &                             &                             &                           &                       &                        & \checkmark                 &                           &                        &                        &                    &                              &                           &                               &                                 &                                 &                                         &                                \\ \hline
				{Confirmation bias}            &                          &                               & \checkmark           &                           &                             &                     &                              &                      &                          &                     &                      &                         &                    &                             &                             &                           &                       &                        &                            &                           &                        &                        &                    &                              &                           &                               &                                 & \checkmark                      & \checkmark                              &                                \\ \hline
				{Conjunction fallacy}          &                          &                               & \checkmark           &                           &                             &                     &                              & \checkmark           &                          &                     &                      &                         &                    &                             &                             &                           &                       &                        &                            &                           &                        &                        &                    &                              &                           &                               &                                 &                                 & \checkmark                              &                                \\ \hline
				{Consistency bias}             &                          &                               &                      &                           &                             &                     &                              &                      &                          &                     &                      &                         &                    &                             &                             &                           &                       &                        &                            &                           &                        &                        &                    &                              & \checkmark                &                               &                                 &                                 &                                         &                                \\ \hline
				{Decoy effect}                 &                          &                               &                      &                           &                             &                     &                              &                      &                          &                     &                      &                         &                    & \checkmark                  &                             &                           &                       &                        &                            &                           &                        &                        &                    &                              &                           &                               &                                 &                                 &                                         &                                \\ \hline
				{Distance effect}              &                          &                               &                      &                           &                             &                     &                              &                      &                          &                     &                      &                         &                    &                             &                             & \checkmark                &                       &                        &                            &                           &                        &                        &                    &                              &                           &                               &                                 &                                 &                                         &                                \\ \hline
				{Egocentric Bias}              &                          &                               &                      &                           &                             &                     &                              &                      &                          &                     &                      & \checkmark              &                    &                             &                             &                           &                       &                        & \checkmark                 &                           &                        &                        &                    &                              &                           &                               &                                 &                                 &                                         &                                \\ \hline
				{Emotional contagion}          &                          &                               &                      &                           &                             &                     &                              &                      & \checkmark               &                     &                      &                         &                    &                             &                             &                           &                       &                        &                            &                           &                        &                        &                    &                              &                           &                               &                                 &                                 &                                         &                                \\ \hline
				{Endowment effect}             &                          &                               &                      &                           &                             &                     &                              & \checkmark           &                          &                     &                      &                         &                    &                             &                             &                           &                       &                        &                            &                           &                        &                        &                    &                              &                           &                               &                                 &                                 & \checkmark                              &                                \\ \hline
				{Framing effect}               &                          &                               & \checkmark           & \checkmark                &                             &                     & \checkmark                   & \checkmark           &                          &                     &                      &                         &                    &                             &                             &                           &                       &                        &                            &                           &                        & \checkmark             &                    &                              &                           &                               &                                 & \checkmark                      &                                         & \checkmark                     \\ \hline
				{In-group bias}                &                          &                               &                      &                           &                             &                     &                              &                      & \checkmark               &                     &                      &                         &                    &                             &                             &                           &                       &                        &                            &                           &                        &                        &                    &                              &                           &                               &                                 &                                 &                                         &                                \\ \hline
				{Insensitivity to sample size} &                          &                               & \checkmark           &                           &                             &                     & \checkmark                   &                      &                          &                     &                      &                         &                    &                             &                             &                           &                       &                        &                            &                           &                        &                        &                    &                              &                           &                               &                                 &                                 & \checkmark                              &                                \\ \hline
				{Knowledge effect}             &                          &                               & \checkmark           &                           & \checkmark                  &                     &                              &                      &                          &                     &                      &                         &                    &                             &                             &                           &                       &                        &                            &                           &                        &                        &                    &                              &                           &                               &                                 &                                 &                                         &                                \\ \hline
				{Odd/even scale effects}       &                          &                               &                      &                           &                             &                     &                              &                      &                          &                     &                      &                         &                    &                             &                             &                           &                       &                        &                            &                           &                        & \checkmark             &                    &                              &                           &                               &                                 &                                 &                                         &                                \\ \hline
				{Opinion floating}             &                          &                               &                      &                           &                             &                     &                              &                      &                          &                     &                      &                         &                    &                             &                             &                           &                       &                        &                            &                           &                        & \checkmark             &                    &                              &                           &                               &                                 &                                 &                                         &                                \\ \hline
				{Order bias}                   &                          &                               &                      &                           &                             &                     &                              &                      &                          & \checkmark          & \checkmark           & \checkmark              & \checkmark         &                             & \checkmark                  &                           & \checkmark            & \checkmark             & \checkmark                 &                           &                        & \checkmark             & \checkmark         & \checkmark                   &                           &                               &                                 &                                 &                                         &                                \\ \hline
				{Overweighting bias}           &                          &                               & \checkmark           &                           &                             &                     &                              &                      &                          &                     &                      &                         &                    &                             &                             &                           &                       &                        &                            &                           &                        &                        &                    &                              &                           &                               &                                 &                                 &                                         &                                \\ \hline
				{Popularity bias}              &                          &                               &                      &                           &                             &                     &                              &                      &                          & \checkmark          &                      &                         &                    &                             &                             &                           &                       &                        &                            &                           &                        &                        &                    &                              &                           &                               &                                 &                                 &                                         &                                \\ \hline
				{Positivity bias}              &                          &                               &                      &                           &                             &                     &                              &                      & \checkmark               &                     &                      &                         &                    &                             &                             &                           &                       &                        &                            &                           &                        &                        &                    &                              &                           &                               &                                 &                                 &                                         &                                \\ \hline
				{Primacy effect}               &                          &                               &                      &                           &                             &                     &                              &                      &                          &                     &                      &                         &                    &                             &                             &                           &                       &                        &                            &                           & \checkmark             &                        &                    &                              &                           & \checkmark                    &                                 &                                 &                                         & \checkmark                     \\ \hline
				{Priming effect}               & \checkmark               & \checkmark                    &                      &                           &                             &                     &                              &                      &                          &                     &                      &                         &                    &                             &                             & \checkmark                &                       &                        &                            &                           &                        &                        &                    &                              &                           &                               &                                 &                                 &                                         &                                \\ \hline
				{Representative bias}          &                          &                               &                      &                           &                             &                     & \checkmark                   &                      &                          &                     &                      &                         &                    &                             &                             &                           &                       &                        &                            &                           &                        &                        &                    &                              &                           &                               &                                 & \checkmark                      & \checkmark                              & \checkmark                     \\ \hline
				{Size congruity effect}        &                          &                               &                      &                           &                             &                     &                              &                      &                          &                     &                      &                         &                    &                             &                             & \checkmark                &                       &                        &                            &                           &                        &                        &                    &                              &                           &                               &                                 &                                 &                                         &                                \\ \hline
				{SNARC effect}                 &                          &                               &                      &                           &                             &                     &                              &                      &                          &                     &                      &                         &                    &                             &                             & \checkmark                &                       &                        &                            &                           &                        &                        &                    &                              &                           &                               &                                 &                                 &                                         &                                \\ \hline
				{Status quo bias}              &                          &                               &                      &                           &                             &                     &                              &                      &                          &                     &                      &                         &                    &                             &                             & \checkmark                &                       &                        &                            &                           &                        &                        &                    &                              &                           &                               &                                 & \checkmark                      &                                         & \checkmark                     \\ \hline
				{Suggestibility}               &                          &                               &                      &                           &                             &                     &                              &                      &                          &                     &                      &                         &                    &                             &                             &                           &                       &                        &                            &                           &                        &                        &                    &                              &                           &                               & \checkmark                      &                                 &                                         &                                \\ \hline
				{Verbosity bias}               &                          &                               &                      &                           &                             &                     &                              &                      &                          &                     &                      & \checkmark              & \checkmark         &                             &                             &                           &                       &                        & \checkmark                 & \checkmark                &                        &                        &                    &                              &                           &                               &                                 &                                 &                                         &                                \\ \hline
				{Von Restorff effect}          &                          &                               & \checkmark           &                           &                             &                     &                              &                      &                          &                     &                      &                         &                    &                             &                             &                           &                       &                        &                            &                           &                        &                        &                    &                              &                           &                               &                                 &                                 &                                         &                                \\ \hline
			\end{tabular}
		}
	\end{table*}
}

The contributions of this study are (i) an extensive survey of existing studies on cognitive biases and their mitigation in LLMs to organize and summarize their results systematically,
(ii) application of two bias mitigation methods used in crowdsourcing to prompts in LLMs, and 
(iii) experimental results showing different effectiveness when used for LLMs than when used in crowdsourcing.

\section{Cognitive biases in LLMs}\label{sec-cognitivebias-in-llms}

Many studies show that LLMs exhibit various cognitive biases that lead to irrational output.
Table \ref{tabl:cb-works} shows which type of cognitive bias each study addresses.

CoBBLEr (Cognitive Bias Benchmark for Large Language Models as Evaluators) \cite{koo2024benchmarking} is a benchmark used to assess the cognitive biases of LLMs.
CoBBLEr addresses six cognitive biases: order bias, compassion fade, egocentric bias, bandwagon effect, attentional bias, and verbosity bias.
Other studies have shown that these biases reduce the performance of LLMs.

\paragraph{Order bias.}
Order bias \cite{israel1990can} is the tendency to prefer an option based on its position in a list rather than the intrinsic quality of its content. Generally, humans favor the option at the top of a list \cite{ayidiya1990response}. Several studies have identified similar imbalances in the outputs of LLMs \cite{hou2024large, zheng2023large, zheng2023judging}. In many cases, LLMs prefer the first option \cite{koo2024benchmarking, pezeshkpour2024large, ranaldi2024hans, tjuatja2024llms, wang2024large, wang2023primacy, wu2023style}.

\paragraph{Compassion fade.}
Compassion fade \cite{butts2019helping} is the decrease in helping or compassionate intent and behavior as the number of people in need increases. LLMs, like humans, behave differently depending on whether the name being evaluated is anonymized or not \cite{koo2024benchmarking}.

\paragraph{Egocentric bias.}
Egocentric bias \cite{ross1979egocentric} is the tendency to rely on one's perspective or to evaluate oneself favorably. Some studies indicate that LLMs prioritize their responses regardless of quality, which can produce irrational results \cite{zheng2023judging, koo2024benchmarking}.

\paragraph{Bandwagon effect.}
Bandwagon effect \cite{leibenstein1950bandwagon} is the tendency to prefer an option simply because it is favored by the majority. LLMs are vulnerable to this bias and prefer the option chosen by the majority, which compromises their decision-making accuracy \cite{koo2024benchmarking}.

\paragraph{Attentional bias.}
Attentional bias is the influence of selective factors on a person's perception \cite{bar-haim2007threatrelated}. We explore one of attentional bias: distraction. Distraction is the tendency for a person’s attention to be diverted and their judgment distorted by irrelevant information. Some studies show that the performance of LLMs is compromised when prompts include irrelevant information \cite{koo2024benchmarking, shi2023large}.

\paragraph{Verbosity bias.}
Verbosity bias \cite{zheng2023judging} is the tendency to favor longer responses, regardless of their quality. This bias is a form of salience bias, where irrational judgments are made by giving undue attention to more noticeable attributes \cite{schenk2011exploiting}. Some studies indicate that LLMs tend to prefer longer responses due to this bias, impacting their effectiveness and accuracy \cite{koo2024benchmarking, zheng2023judging, wu2023style, saito2023verbosity}.

\section{Mitigating cognitive biases of LLMs}\label{sec-mitigating-cognitivebias}

\subsection{Existing mitigation methods for LLMs}\label{subsec-existing-methods}
Some Techniques are proposed to mitigate cognitive bias in LLMs.

Several methods exist to address order bias and primacy effect in multiple-choice questions: inputting the same question multiple times while shuffling the positions of the options \cite{hou2024large, pezeshkpour2024large, wang2024large, zheng2023judging}, learning additional parameters \cite{zhao2021calibrate}, Bayesian probabilistic framework \cite{ma2023large}, self-supervised position debiasing framework \cite{liu2024selfsupervised}, controlling the output so that it follows a particular format or structure \cite{eicher2024reducing}. There is also a method to remove the bias against the labels of the options \cite{zheng2023large}. However, these methods are specific to order bias and are not used for other types of cognitive bias.

For attentional bias, there is a method to prompt LLMs to ignore irrelevant information to help them focus on the essential aspects of tasks \cite{shi2023large}.
However, this method is not used for other biases. 

Across different types of cognitive biases, some general methods have been developed to improve the outputs of LLMs.
For example, there is a method to force LLMs to provide the reason before LLMs output the evaluation \cite{wang2024large}.
These biases can also be mitigated by using techniques to devise prompts, such as those used in general tasks: few-shot prompting \cite{brown2020language} with well-balanced examples \cite{zheng2023judging, ranaldi2024hans}, Chain-of-Thought Prompting \cite{wei2022chainthought}, Zero-shot Chain-of-Thought Prompting \cite{kojima2022large}, a method of giving several examples of errors and having them corrected \cite{schmidgall2024addressing}, and changing prompts by the LLM itself \cite{echterhoff2024cognitive}.
While these strategies can be used for a broad range of cognitive biases, their inputs and outputs are longer than necessary or the same problem needs to be solved many times.

\subsection{Applying cognitive bias mitigation methods in crowdsourcing to prompts of LLMs}\label{sec-proposedmethod}
There are methods for mitigating cognitive biases in crowdsourcing.
Cognitive biases can significantly degrade the quality of annotations obtained through crowdsourcing \cite{eickhoff2018cognitive}.
In order to mitigate biases in crowdsourcing, there are two techniques for making changes to the instructions \cite{hube2019understanding}: SoPro and AwaRe.
SoPro asks crowdworkers to predict the label that they believe the majority of other workers would choose.
AwaRe makes crowdworkers aware of their inherent biases before they answer.
These two methods can also be used for cognitive biases \cite{hettiachchi2021investigating}.

Therefore, inspired by these studies, we adapted these mitigation methods in crowdsourcing to the prompts of LLMs.

\paragraph{SoPro:}\label{par-sopro}
SoPro (Social Projection) adds a sentence to the prompt instructing LLMs to consider how the majority of people would respond. For example, the following statement is added at the beginning: ``Please answer the following question according to how you believe the majority of people would answer.'' This method encourages LLMs to reflect broader social consensus rather than individual or idiosyncratic interpretations.

\paragraph{AwaRe:}\label{par-aware}
AwaRe (Awareness Reminder) adds a statement that informs the presence of a bias and instructs LLMs to be careful of this bias while answering. In addressing order bias, for example, the following statement is added at the beginning: ``Please answer the following question while being aware of order bias.'' This method is designed to prompt LLMs to consciously adjust their responses to mitigate the influence of the bias.

These methods have two advantages over existing methods.
First, there is no need to output an explanation of the answer, and the same problem is not solved repeatedly, resulting in less lengthy inputs and outputs.
Second, our methods are versatile and can be applied to any cognitive bias, as they are not limited by question format or type of bias.

\section{Experiments}\label{sec-experiment}

To evaluate the effectiveness of our methods, we conducted experiments using the CoBBLEr benchmark.
The CoBBLEr benchmark \cite{koo2024benchmarking} quantifies the influence of six cognitive biases on evaluations of text quality by LLMs.
In this experiment, we applied this benchmark to both GPT-3.5 and GPT-4 in a manner consistent with Koo et al. \cite{koo2024benchmarking}.
Details of the LLMs used for this experiment are provided in Appendix \ref{app:LLM-detail}. 
Results obtained without using any of the methods are used as a baseline.
In addition, the existing method of asking LLMs the reason for their answer \cite{wang2024large} was used for comparison.

\subsection{Evaluation method}\label{subsec-cob-setting}
The six biases addressed in this experiment are order bias, compassion fade, egocentric bias, bandwagon effect, attentional bias, and verbosity bias.
We conducted experiments to assess the influence of each of these biases.

CoBBLEr consists of 50 question-answer pairs and outputs obtained by inputting each question into 16 different LLMs.
These pairs are used to evaluate an LLM by having it perform the following tasks.
For each question, a pair of two answers is chosen from the 16 responses, and the LLM evaluates which of the two is more consistent.
This evaluation is conducted for all pairs of 16 LLMs and then for all 50 questions.

To assess whether the LLM is affected by bias, one pair is evaluated twice with change.
For example, the LLM evaluates once and then evaluates again in the order of the choices rearranged.
These evaluations are then examined to determine if they are affected by order bias.
In total, the LLM performs 12,000 evaluations per bias.
Except for compassion fade, the names of LLMs are anonymized by randomly replacing responses and changing names to ``\verb|System Star|'' and ``\verb|System Square|'' so as not to affect the evaluation process.

\subsection{Evaluation setting}
We describe how the prompts are modified to account for each bias, and then define the scores used to assess the influence of the bias. This modification of the prompts is consistent with that of Koo et al. The modified prompt templates are shown in Appendix \ref{app:prompt-template}. Let $Y=0$ denote that the first response is better, and $Y=1$ denote that the second response is better. Let $P$ denote the ratio of all response pairs.

\paragraph{Order bias. }
To address order bias, we input two prompts to LLMs when evaluating responses: one with the first response listed first, and the other with the second response listed first. If order bias does not influence the evaluation, the LLM should select the same response in both cases. We consider it an inconsistent response when different outputs are chosen.

Let $A=0$ indicate that the first answer is presented first and $A=1$ indicate that the second answer is presented first. We define \soone \ as the score when the first answer is chosen in both cases and \sotwo \ as the score when the second answer is chosen in both cases, calculated as follows:
\begin{align}
 \soone & = P\left((Y=0, A=0) \land (Y=1, A=1)\right), \\
 \sotwo & = P\left((Y=1, A=0) \land (Y=0, A=1)\right).
\end{align}
If the LLM is not affected by order bias, both \soone \ and \sotwo \ should be 0.

\paragraph{Compassion fade. }
We examine the outputs of LLMs when the actual model names are used instead of anonymized names. When investigating compassion fade, the analysis is complicated by the simultaneous influence of order bias. Therefore, we perform the same comparisons as in the study of order bias, using both anonymized and actual model names, and then compare the results. If the LLM is unaffected by compassion fade, it should output consistent results regardless of whether the model names are anonymized or not.

Let $A=0$ indicate that the first answer is presented first, and $A=1$ indicate that the second answer is presented first. Additionally, let $Y^\prime=0$ when the LLM determines that the first response is better, and $Y^\prime=1$ when the LLM determines that the second response is better. The scores are defined as follows:
\begin{align}
 \scone & = P\left((Y^\prime=0, A=0) \land (Y^\prime=0, A=1)\right), \\
 \sctwo & = P\left((Y^\prime=1, A=0) \land (Y^\prime=1, A=1)\right).
\end{align}
If the LLM is not affected by compassion fade, \scone \ and \sctwo \ should be equal to \soone \ and \sotwo, respectively.

\paragraph{Egocentric bias. }
When evaluating responses, we anonymize LLMs and indicate one of the responses as the LLM's own by adding ``\verb|(You)|'' to either ``\verb|System Star|'' or ``\verb|System Square|''. If the LLM is free from this bias, it should select the same answer regardless of the presence of ``\verb|(You)|''. To test this, we modify the responses in two ways: adding ``\verb|(You)|'' to the first response and then to the second. If the LLM consistently selects the response containing ``\verb|(You)|'', it suggests that the LLM is influenced by this bias.

We define $A=0$ when ``\verb|(You)|'' is added to the first response and $A=1$ when it is added to the second response. The score \sego \ for evaluating the effect of egocentric bias is defined as follows:
\begin{equation}
 \sego = P\left((Y=0, A=0) \land (Y=1, A=1)\right).
\end{equation}
If the LLM is not affected by egocentric bias, \sego \ should be 0.

\paragraph{Bandwagon effect. }
To investigate bandwagon effect, we modify the prompts by adding a statement indicating that the majority of people prefer either the first or the second response, like ``80\% of people believe that System Star is better.'' If the LLM selects different responses based on this indication, its responses are considered inconsistent, suggesting susceptibility to bandwagon effect.

We define $A=0$ when the statement indicates that the first response is preferred, and $A=1$ when the second response is indicated as preferred. The score \sband \ for evaluating the impact of bandwagon effect is defined as follows:
\begin{equation}
 \sband = P\left((Y=0, A=0) \land (Y=1, A=1)\right).
\end{equation}
If the LLM is not affected by bandwagon effect, \sband \ should be 0.

\paragraph{Attentional bias. }
To assess attentional bias, we add irrelevant information from either the first or the second response. For instance, we append to the instructions that ``\verb|System Star| likes to eat apples and oranges.'' If the LLM is not affected by attentional bias, they should choose the same response regardless of the presence of the information. If the LLM selects the response containing irrelevant information in both cases, it is indicative that the LLM is exhibiting inconsistent responses influenced by attentional bias.

We set $A=0$ when irrelevant information is added to the first response and $A=1$ when it is added to the second response. The score \sattn \ for evaluating the impact of attentional bias is defined as follows:
\begin{equation}
 \sattn = P\left((Y=0, A=0) \land (Y=1, A=1)\right).
\end{equation}
If the LLM is not affected by attentional bias, \sattn \ should be 0.

\paragraph{Verbosity bias. }
When the LLM evaluates the responses, we compare the percentage of longer responses it selects, based on the number of tokens in the responses.
Without the influence of verbosity bias, the percentage of longer responses should be the same as that of shorter responses.

Let $A=0$ be when the first response has more tokens and $A=1$ when the second response has more tokens.
In this case, the score \ssal \ for evaluating the effect of verbosity bias is as follows:
\begin{equation}
 \ssal = P\qty(Y=0, A=0) + P\qty(Y=1, A=1) - 0.5.
\end{equation}
If the LLM is not affected by verbosity bias, \ssal \ should be 0.

\subsection{Results}\label{subsubsec-cob-result}
The results of this experiment are presented in Table \ref{tabl:cobbler} and Table \ref{tabl:attentionbias}. We use the results from inputting a modified prompt for each bias as the baseline and compare these with the outcomes from applying the existing method \cite{wang2024large} and our methods to this prompt. In these tables, higher scores are colored red, and lower scores are colored blue, relative to the case where the responses are random. Furthermore, the intensity of the color increases with the deviation from the random response case. This color coding indicates that red scores signify less consistency and greater influence by bias, whereas blue scores suggest greater consistency and less influence by bias. In this section, we analyze the results by each bias.

\begin{table*}[tbp]
 \centering
 \caption{Comparison of scores for all methods. ``With reason'' is the existing method. ``SoPro'' and ``AwaRe'' are our methods. Compared to random responses, strengthened bias is indicated in red, and weakened bias in blue. For compassion fade, the closer \scone ~and \sctwo ~are to the \soone ~and \sotwo, respectively, the more favorable the results. For the other biases, the closer the score is to 0, the more favorable the result.}
 \label{tabl:cobbler}
 \begin{tabular}{c|c|cc|cc|c|c|c|c}\hline
                              &        & \multicolumn{2}{c|}{Order}   & \multicolumn{2}{c|}{Comp.}   & Egoc.                       & Band.                       & Attn.                       & Verb.                                                                                   \\
                             LLM& Method     & \soone                      & \sotwo                      & \scone                      & \sctwo                      & \sego                       & \sband                      & \sattn                      & \ssal                      \\\hline \hline
                            - & Random   & 0.25                        & 0.25                        & 0.25                        & 0.25                        & 0.25                        & 0.25                        & 0.25                        & 0.0                        \\\hline
   \multirow{4}{*}{GPT-3.5} & Baseline & \colorbox{blue!5.0}{0.200}  & \colorbox{blue!20.8}{0.042} & \colorbox{blue!18.3}{0.067} & \colorbox{blue!11.3}{0.137} & \colorbox{blue!17.4}{0.076} & \colorbox{red!27.4}{0.524}  & \colorbox{blue!24.9}{0.001} & \colorbox{red!9.6}{0.096}  \\
                            & With reason   & \colorbox{blue!3.9}{0.211}  & \colorbox{blue!21.0}{0.040} & \colorbox{blue!18.9}{0.061} & \colorbox{blue!7.9}{0.171}  & \colorbox{blue!16.0}{0.090} & \colorbox{red!11.8}{0.368}  & \colorbox{blue!24.9}{0.001} & \colorbox{red!10.8}{0.108} \\
                            & SoPro  & \colorbox{blue!2.7}{0.223}  & \colorbox{blue!21.5}{0.035} & \colorbox{blue!17.1}{0.079} & \colorbox{blue!9.7}{0.153}  & \colorbox{blue!23.6}{0.014} & \colorbox{red!70.5}{0.955}  & \colorbox{blue!24.9}{0.001} & \colorbox{red!9.6}{0.096}  \\
                            & AwaRe  & \colorbox{blue!5.9}{0.191}  & \colorbox{blue!20.5}{0.045} & \colorbox{blue!14.7}{0.103} & \colorbox{blue!13.5}{0.115} & \colorbox{blue!19.8}{0.052} & \colorbox{red!1.0}{0.260}   & \colorbox{blue!25}{0.000}   & \colorbox{red!11.9}{0.119} \\\hline
   \multirow{4}{*}{GPT-4}   & Baseline & \colorbox{blue!18.8}{0.062} & \colorbox{blue!16.0}{0.090} & \colorbox{blue!18.8}{0.062} & \colorbox{blue!17.7}{0.073} & \colorbox{blue!22.5}{0.025} & \colorbox{blue!3.6}{0.214}  & \colorbox{blue!24.3}{0.007} & \colorbox{red!3.1}{0.031}  \\
                            & With reason & \colorbox{blue!19.1}{0.059} & \colorbox{blue!16.7}{0.083} & \colorbox{blue!15.7}{0.093} & \colorbox{blue!19.8}{0.052} & \colorbox{blue!21.6}{0.034} & \colorbox{blue!8.5}{0.165}  & \colorbox{blue!24.0}{0.010} & \colorbox{red!3.8}{0.038}  \\
                            & SoPro  & \colorbox{blue!19.6}{0.054} & \colorbox{blue!15.3}{0.097} & \colorbox{blue!18}{0.070}   & \colorbox{blue!18.3}{0.067} & \colorbox{blue!21.3}{0.037} & \colorbox{red!0.5}{0.255}   & \colorbox{blue!24.2}{0.008} & \colorbox{red!2.4}{0.024}  \\
                            & AwaRe  & \colorbox{blue!20.7}{0.043} & \colorbox{blue!14.0}{0.110} & \colorbox{blue!18.6}{0.064} & \colorbox{blue!17.4}{0.076} & \colorbox{blue!22.3}{0.027} & \colorbox{blue!10.8}{0.142} & \colorbox{blue!24.3}{0.007} & \colorbox{red!1.1}{-0.011} \\\hline
 \end{tabular}
\end{table*}

\begin{table*}[tbp]
	\centering
    \caption{Comparison of the scores for attentional bias and the percentage of valid responses before and after correction. In the correction, responses to either option are considered ``valid'' even if the format is invalid. ``With reason'' is the existing method. ``SoPro'' and ``AwaRe'' are our methods. Compared to random responses, enhanced bias is highlighted in red, and weakened bias in blue. The closer \sattn and percentage of valid responses are to 0 and 1, respectively, the more favorable the results.}
	\label{tabl:attentionbias}
	\begin{tabular}{c|c|cc|cc}\hline
		                                  &  & \multicolumn{2}{c|}{Before}     & \multicolumn{2}{c}{After}                                                              \\
		                                 LLM & Method & \sattn                      & Valid                       & \sattn                      & Valid                    \\\hline \hline
		                          - &Random   & 0.25                        & 1.000                      & 0.25                        & 1.000                     \\\hline
		\multirow{4}{*}{GPT-3.5}   & Baseline & \colorbox{blue!24.9}{0.001} & \colorbox{red!28.7}{0.713} & \colorbox{blue!24.6}{0.004} & 1.000                     \\
		                         & With reason & \colorbox{blue!24.9}{0.001} & 1.000                      & \colorbox{blue!24.9}{0.001} & 1.000                     \\
		                         & SoPro  & \colorbox{blue!24.9}{0.001} & \colorbox{red!52.8}{0.472} & \colorbox{blue!24.6}{0.004} & \colorbox{red!0.1}{0.999} \\
		                         & AwaRe  & \colorbox{blue!25.0}{0.000} & \colorbox{red!12.7}{0.873} & \colorbox{blue!24.9}{0.001} & 1.000                     \\\hline
		\multirow{4}{*}{GPT-4}   & Baseline & \colorbox{blue!24.3}{0.007} & \colorbox{red!0.5}{0.995}  & \colorbox{blue!24.3}{0.007} & \colorbox{red!0.5}{0.995} \\
		                         & With reason & \colorbox{blue!24.0}{0.010} & \colorbox{red!0.2}{0.998}  & \colorbox{blue!24.0}{0.010} & \colorbox{red!0.2}{0.998} \\
		                         & SoPro  & \colorbox{blue!24.2}{0.008}         & \colorbox{red!0.5}{0.995}         & \colorbox{blue!24.2}{0.008}         & \colorbox{red!0.5}{0.995}        \\
		                         & AwaRe  & \colorbox{blue!24.3}{0.007} & \colorbox{red!0.2}{0.998}  & \colorbox{blue!24.3}{0.007} & \colorbox{red!0.2}{0.998} \\\hline
	\end{tabular}
\end{table*}

\paragraph{Order bias.}
At baseline, GPT-3.5 shows a preference for the first option, while GPT-4 tends to prefer the second option. Both GPT-3.5 and GPT-4 are susceptible to order bias, but the effect is smaller for GPT-4 than for GPT-3.5. Despite the application of the existing method and our methods, no significant changes in performance are observed for either GPT-3.5 or GPT-4 relative to their baselines.

\paragraph{Compassion fade.}
Since the evaluation of compassion fade may be influenced by order bias, we analyze its effect by comparing the result with that of order bias. At baseline, GPT-3.5 prefers the last option, contrasting with its behavior when the model name is anonymized during the evaluation of order bias, thus demonstrating the influence of compassion fade. This influence is less significant in GPT-4 than in GPT-3.5.
For GPT-3.5, the tendency is more marked when the existing method or SoPro is applied, and slightly reduced when AwaRe is employed. For GPT-4, there are no significant changes.

\paragraph{Egocentric bias.}
At baseline, GPT-3.5 and GPT-4 prefer their responses although this effect is smaller than when responses are chosen at random. For GPT-3.5, this tendency slightly increases when the existing method is applied, but is somewhat reduced when SoPro or AwaRe is applied. For GPT-4, no significant changes are observed.

\paragraph{Bandwagon effect.}
At baseline, both models exhibit a preference for the response selected by the majority. This tendency is more likely in GPT-3.5 than in GPT-4. The most effective mitigation across both GPT-3.5 and GPT-4 is observed when AwaRe is applied. Conversely, SoPro significantly worsens the performance.

\paragraph{Attentional bias.}
Table \ref{tabl:cobbler} shows that the effect of attentional bias is minimal in the baseline, and neither the existing method nor our methods significantly alter these results. 
However, GPT-3.5 makes a high proportion of invalid responses that do not comply with the prescribed format. 
Table \ref{tabl:attentionbias} presents the scores and percentages of valid responses before and after correcting invalid outputs. 
For these analyses, a response is considered valid if it indicates which answer is better, regardless of adherence to the format. 
The results of the baseline indicate a significant number of responses are invalid, which can be attributed to attentional bias. 
The application of the existing method and AwaRe results in more robust outputs, whereas the performance of SoPro falls below the baseline. 
After corrections, the effect of attentional bias remains minimal at baseline and exhibits no change when the mitigation methods are applied. 
In contrast, the majority of responses from GPT-4 at baseline are considered valid, indicating a lesser impact of attentional bias on this model.

\paragraph{Verbosity bias.}
At baseline, GPT-3.5 prefers longer responses due to verbosity bias, which is less shown in GPT-4. When applying both the existing method and our methods, no one succeeds in improving the performance of GPT-3.5. Conversely, GPT-4 shows improvement through the application of SoPro and AwaRe.

\subsection{Discussion}
GPT-4 is generally less susceptible to cognitive biases compared to GPT-3.5. At baseline, GPT-3.5 exhibits robustness to order bias, compassion fade, and egocentric bias. However, it is significantly affected by bandwagon effect, attentional bias, and verbosity bias, demonstrating a lack of robustness against these biases. Conversely, GPT-4 shows a reduced susceptibility to all six biases. The effects of bandwagon effect and verbosity bias are present but less marked than in GPT-3.5, indicating higher resistance to these biases.

The existing method proves effective for mitigating bandwagon effect and attentional bias in GPT-3.5, as well as bandwagon effect in GPT-4. This effectiveness can be attributed to the method's prompting of LLMs to provide reasons for their answers, which encourages more rational responses. This aligns with findings from previous studies. In the case of attentional bias, there is a significant increase in the percentage of valid responses for GPT-3.5, indicating success in mitigating this bias.

SoPro is effective in mitigating egocentric bias in GPT-3.5 and verbosity bias in GPT-4. This effectiveness may be attributed to the more objective responses elicited by having LLMs respond based on how they think the majority of people would respond. However, SoPro increases the models' susceptibility to bandwagon effect, as it inherently aligns their responses with the opinion of the majority. Although SoPro is designed to enhance sensitivity to social perspectives, its effectiveness is limited for cognitive biases where conformity to group norms is less desirable. These results are inconsistent with the claims of the study when used on humans.

AwaRe is effective on GPT-3.5 against order bias, compassion fade, egocentric bias, bandwagon effect, and attentional bias. Furthermore, it mitigated the effects of the bandwagon effect and verbosity bias on GPT-4. These results suggest that AwaRe makes LLMs more aware of their biases and allows them to make careful judgments, thereby enabling them to respond more rationally.

\section{Related work}\label{sec-related-works}

\subsection{Analysis of cognitive bias in LLMs}
Many studies have demonstrated that LLMs are susceptible to cognitive biases.
Table \ref{tabl:cb-works} shows which type of cognitive bias each study addresses.

Several studies have highlighted that LLMs display cognitive biases similar to humans.
Suri et al. \cite{suri2024large} investigate whether GPT-3.5 employs human-like decision heuristics and biases, such as anchoring, and find that GPT-3.5 exhibits similar effects to humans in various tests.
Lampinen et al. \cite{lampinen2024language} demonstrate that LLMs show knowledge effects in reasoning, where performance on logical tasks improves when semantic content aligns with correct logical inferences.
Shaki et al. \cite{shaki2023cognitive} confirmed a range of cognitive biases, like priming effect, in GPT-3.

LLMs also respond similarly to human social cognitive patterns.
Bian et al. \cite{bian2024influence} explore how external statements and opinions influence the cognition and behaviors of LLMs, discovering that their responses to external information reflect human social cognitive patterns, such as authority and in-group biases.

Cognitive biases that differ from those in humans may reduce LLM's ability to reason.
Macmillan-Scott et al. \cite{macmillan-scott2024irrationality} evaluated LLMs on a cognitive psychology task to determine whether they make rational answers for mathematical questions, and showed that LLMs exhibit irrational biases that are distinct from humans. 
Opedal et al. \cite{opedal2024language} examined whether LLMs' responses include the cognitive biases that children exhibit when solving problems, and found that LLMs exhibited different biases from children.

Several studies have concentrated on the types of LLMs and their specific cognitive biases. 
Hagendorff et al. \cite{hagendorff2023thinking} report that while earlier models like GPT-3 exhibit human-like intuitive behaviors and cognitive errors, more advanced models, such as ChatGPT and GPT-4, show improvements, overcoming these errors and demonstrating rational decision-making as evidenced by psychology-based tests like the Cognitive Reflection Test.
Itzhak et al. \cite{itzhak2024instructed} explore the impact of instruction tuning and RLHF (Reinforcement Learning from Human Feedback) \cite{ouyang2022training} on decision-making in LLMs, particularly analyzing the presence of decoy effect, certainty effect, and belief bias in instruction-tuned models such as GPT-3.5.
Tjuatja et al. \cite{tjuatja2024llms} investigate human-like response biases in survey design by LLMs, finding that popular models, especially after RLHF, generally do not accurately mimic human behavior.
These are consistent with the result of Casper et al. that point out significant problems and inherent flaws in RLHF \cite{casper2023open}.

Some studies are exploring the effects of its application in specific areas.
Jones and Steinhardt \cite{jones2022capturing} focus primarily on code generation models, demonstrating that some biases such as framing can degrade the quality of their outputs.
Since cognitive bias in medical LLM can lead to inaccurate diagnosis and treatment recommendations, the BiasMedQA dataset was introduced to assess this \cite{schmidgall2024addressing}.
Several studies have addressed order and selection bias, which affect the reliability of recommendation systems \cite{ma2023large, hou2024large}.
Talboy and Fuller \cite{talboy2023challenging} examine the impact of several biases on LLMs and advocate for enhanced education, risk management, and best practices to ensure responsible adoption of this technology.

There is a study working on hallucinations by understanding cognitive biases.
Berberette et al. \cite{berberette2024redefining} classify hallucination by cognitive biases such as the availability heuristic and psychological concepts such as cognitive dissonance.

\subsection{Comparison of LLMs and humans}
Many studies investigate whether LLMs can serve as a viable alternative to humans in various NLP tasks.
Chiang and Lee \cite{chiang2023can} demonstrate that LLMs are comparable to human experts in assessing text quality.
Wang et al. \cite{wang2023chatgpt} indicate that ChatGPT demonstrates a high correlation with human judgment when used as a natural language generation evaluation metric.
Furthermore, it is very difficult to distinguish between the responses of humans and LLMs \cite{brown2020language}.

Additionally, several studies demonstrate the performance of LLMs in crowdsourcing annotation tasks.
Gilardi et al. \cite{gilardi2023chatgpt} demonstrate that ChatGPT surpasses crowd workers in various text annotation tasks, achieving higher accuracy at lower cost.
T{\"o}rnberg \cite{tornberg2023chatgpt4} demonstrates that GPT-4 outperforms experts and crowd workers in accurately and reliably classifying the political affiliation of text from U.S. politicians, with comparable or reduced bias.

Some studies examine whether LLMs replicate findings established in social sciences, such as economics, psycholinguistics, and social psychology.
Aher et al. \cite{aher2023using} evaluate the ability of LLMs to simulate diverse human behaviors and show that ChatGPT and GPT-4 are effective at replicating human behavior and answering questions.
Binz and Schulz \cite{binz2023using} conduct cognitive psychology experiments to assess GPT-3’s decision-making and reasoning, revealing vulnerabilities such as poor causal reasoning and sensitivity to task perturbations.
Horton \cite{horton2023large} discusses the capacity of LLMs to serve as computational analogs to humans and to demonstrate how LLMs are appropriate for simulating human behavior.
Sinclair et al. \cite{sinclair2022structural} investigated how structural priming affects LLMs’ ability to learn and utilize abstract structural information.

Other studies investigate how LLMs replicate public opinion.
Argyle et al. \cite{argyle2023out} demonstrate that GPT-3 can accurately reflect diverse human attitudes and serve as a powerful tool for studying human society.
However, Santurkar et al. \cite{santurkar2023whose} demonstrate a significant misalignment of LLMs with public opinion across diverse U.S. demographics and identify demographic groups whose views are underrepresented by LLMs.

\section{Conclusion}\label{sec-conclusion}
In this study, we first surveyed existing studies that examine cognitive biases in LLMs and methods to mitigate them (Table \ref{tabl:cb-works}).
Many studies show that LLMs are affected by various types of cognitive biases.
On the other hand, existing mitigation methods have the disadvantage that they are limited in the type of biases to apply or have lengthy inputs or outputs.

We then addressed the introduction of two mitigation methods adapted from crowdsourcing, SoPro and AwaRe, into the prompts of LLMs.
SoPro encourages LLMs to respond based on how they think the majority would answer.
AwaRe enhances awareness of biases and prompts LLMs to answer with greater attention to mitigating these biases.
Finally, we conducted experiments using the CoBBLEr benchmark to compare the effectiveness of these methods with the existing method. 
The results indicate that GPT-3.5 naturally exhibits robustness to some biases, and GPT-4 demonstrates greater resistance to all tested biases.
The results also revealed that while SoPro was less effective, AwaRe successfully promoted rational responses and mitigated bias effects.

This study has limitations. 
Our focus was primarily on prompt modification as a means of bias mitigation, without exploring alternative approaches like fine-tuning with additional data.
It is possible that these approaches can mitigate the biases.
In addition, AwaRe requires inputting the name of the bias to be mitigated, so it is necessary to know which bias will affect the LLM in advance.
Moreover, the experiments used only GPT-3.5 and GPT-4, and did not address variations in model tuning and training such as instruction tuning and RLHF, or differences in model sizes.
It would be essential to expand this research to include a variety of LLMs with different architectures and training paradigms to fully understand the scope and limitations of various bias mitigation methods.
Furthermore, exploring additional dimensions such as cultural and linguistic variations in data could offer deeper insights into the resilience of LLMs against various cognitive biases. 

\begin{acks}
This work was supported by JST CREST Grant Number JPMJCR21D1.
\end{acks}

\bibliographystyle{ACM-Reference-Format}
\bibliography{fullbibfile}

\newpage
\renewcommand{\thesection}{\Alph{section}}
\setcounter{section}{0}

\section{LLM details}\label{app:LLM-detail}
The version of GPT-3.5 we used is \verb|gpt-3.5-turbo-1106|, and the version of GPT-4 we used is \verb|gpt-4-1106-preview|.
We accessed both models through OpenAI API.
We set the same hyperparameters for both models as follows.
We set \verb|temperature| to 0 and \verb|seed| to 0.

\section{Prompt templates}\label{app:prompt-template}
We use the same template of the prompts used by Koo et al. in our experiment.
These templates are modified from the original prompt according to each bias as the bias affects LLMs.

The instruction is represented by ``\{instruction\},'', the reference by ``\{reference\},'', the response of the first model by ``\{response\_star\},'' and the response of the second model by ``\{response\_square\}.''

\subsection{Original prompt}\label{app-subsec-original}
\begin{screen}
\#\#\# Instruction: Which system's response is more coherent considering the reference and instruction?\\

The instruction is: \{instruction\}\\
The reference is: \{reference\}\\

```\\
System Star: \{response\_star\}\\
System Square: \{response\_square\}\\
```\\

Please respond in the following format strictly:\\

System \_ is better\\

Do not provide any additional text or explanation:\\

\#\#\# Response:
\end{screen}

\subsection{Order bias}\label{app-subsec-orderbias}
\begin{screen}
\#\#\# Instruction: Which system's response is more coherent considering the reference and instruction?\\

The instruction is: \{instruction\}\\
The reference is: \{reference\}\\

```\\
System Square: \{response\_square\}\\
System Star: \{response\_star\}\\
```\\

Please respond in the following format strictly:\\

System \_ is better\\

Do not provide any additional text or explanation:\\

\#\#\# Response:
\end{screen}

\subsection{Compassion fade}\label{app-subsec-compassionfade}
The name of the first model is represented by ``\{model0\_name\},'' the name of the second model by ``\{model1\_name\},'' the response of the first model by ``\{model0\_response\},'' the response of the second model by ``\{model1\_response\}.''
\begin{screen}
\#\#\# Instruction: Which model's response is more coherent considering the reference and instruction?\\

The instruction is: \{instruction\}\\
The reference is: \{reference\}\\

```\\
Model \{model0\_name\}: \{model0\_response\}\\
Model \{model1\_name\}: \{model1\_response\}\\
```\\

Please respond in the following format strictly:\\

Model \_ is better\\

Do not provide any additional text or explanation:\\

\#\#\# Response:
\end{screen}

\subsection{Egocentric bias}\label{app-subsec-egocentricbias}
``(You)'' is added after either ``System Star'' or ``System Square''.
\begin{screen}
\#\#\# Instruction: Which system's response is more coherent considering the reference and instruction?\\

The instruction is: \{instruction\}\\
The reference is: \{reference\}\\

```\\
System Star (You): \{response0\}\\
System Square: \{response1\}\\
```\\

Please respond in the following format strictly:\\

System \_ is better\\

Do not provide any additional text or explanation:\\

\#\#\# Response:
\end{screen}

\subsection{Bandwagon effect}\label{app-subsec-bandwagoneffect}
``\{bandwagon\_percent\}'' is a random value added to affect bandwagon effect and can take on an integer value between 60 and 90.
``\{system\}'' is either ``System Star'' or ``System Square.''
\begin{screen}
\#\#\# Instruction: Which system's response is more coherent considering the reference and instruction?\\

The instruction is: \{instruction\}\\
The reference is: \{reference\}\\

```\\
System Star: \{response0\}\\
System Square: \{response1\}\\
```\\

\{bandwagon\_percent\}\% of people believe that \{system\} is better.\\

Please respond in the following format strictly:\\

System \_ is better\\

Do not provide any additional text or explanation:\\

\#\#\# Response:
\end{screen}

\subsection{Attentional bias}\label{app-subsec-attentionbias}
``\{distraction\}'' is an irrelevant sentence added to affect attentional bias, which can be one of the following four sentences.
``\{system\}'' is either ``System Star'' or ``System Square.''
\begin{itemize}
	\item ``\{system\} likes to eat apples and oranges''
	\item ``The coolest thing that \{system\} can do is a 60 second handstand''
	\item ``\{system\} plays a lot of soccer and basketball''
	\item ``\{system\} has been all around Europe two times''
\end{itemize}
\begin{screen}
\#\#\# Instruction: Which system's response is more coherent considering the reference and instruction?\\

The instruction is: \{instruction\}\\
The reference is: \{reference\}\\

```\\
System Star: \{response0\}\\
System Square: \{response1\}\\
```\\

\{distraction\}.
Please respond in the following format strictly:\\

System \_ is better\\

Do not provide any additional text or explanation:\\

\#\#\# Response:
\end{screen}

\end{document}